\RequirePackage[svgnames]{xcolor}
\documentclass[11pt,letterpaper]{mystyle}

\usepackage[all]{hypcap}
\usepackage[svgnames]{xcolor}
\usepackage[numbers,sort&compress]{natbib}
\usepackage{hyperref}[citecolor=lightblue]

\hypersetup{
    colorlinks = true,
    citecolor = {DodgerBlue},
}
\usepackage{multicol}
\usepackage{setspace}
\usepackage{caption}
\usepackage{ragged2e}
\newcommand{\x}{\mathbf{x}}

\definecolor{blanchedalmond}{rgb}{1.0, 0.92, 0.8}
\definecolor{carmine}{rgb}{0.59, 0.0, 0.09}
\definecolor{lightblue}{rgb}{0.22,0.45,0.70}%

\renewcommand{\mathbf}{\boldsymbol}

\makeatletter
\def\Ddots{\mathinner{\mkern1mu\raise\p@
\vbox{\kern7\p@\hbox{.}}\mkern2mu
\raise4\p@\hbox{.}\mkern2mu\raise7\p@\hbox{.}\mkern1mu}}
\makeatother

\definecolor{amaranth}{rgb}{0.9, 0.17, 0.31}
\definecolor{antiquebrass}{rgb}{0.8, 0.58, 0.46}
\definecolor{antiquefuchsia}{rgb}{0.57, 0.36, 0.51}
\definecolor{chromeyellow}{rgb}{0.31, 0.47, 0.26}

\newcommand{\github}{\raisebox{-1.5pt}{\includegraphics[height=1.05em]{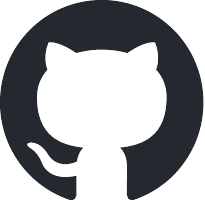}}}

\usepackage{amssymb,mathrsfs,amsmath}

\usepackage[utf8]{inputenc} 
\usepackage[T1]{fontenc}    
\usepackage{url}            
\usepackage{booktabs}       
\usepackage{amsfonts}       
\usepackage{nicefrac}       
\usepackage{microtype}      
\usepackage{array}
\usepackage{graphicx}
\usepackage{wrapfig}
\usepackage{bxcoloremoji}
\usepackage{pifont}
\usepackage{subcaption}
\usepackage{enumitem}
\usepackage[skip=12pt plus 2pt minus 1pt]{parskip}

\usepackage{booktabs}
\usepackage{multirow}
\usepackage{makecell} 
\usepackage{xcolor} 
\usepackage{tabularx}

\definecolor{lightgraycustom}{gray}{0.8}
\usepackage{tcolorbox}

\title{MiroThinker-1.7 \& H1: Towards Heavy-Duty Research Agents via Verification}

\author{
  MiroMind Team \\
}
\runningtitle{MiroThinker-1.7 \& H1 Technical Report}

\begin{document}

\begin{abstract}

We present MiroThinker-1.7, a new research agent designed for complex long-horizon reasoning tasks. Building on this foundation, we further introduce MiroThinker-H1, which extends the agent with heavy-duty reasoning capabilities for more reliable multi-step problem solving.
In particular, MiroThinker-1.7 improves the reliability of each interaction step through an agentic mid-training stage that emphasizes structured planning, contextual reasoning, and tool interaction. This enables more effective multi-step interaction and sustained reasoning across complex tasks.
MiroThinker-H1 further incorporates verification directly into the reasoning process at both local and global levels. Intermediate reasoning decisions can be evaluated and refined during inference, while the overall reasoning trajectory is audited to ensure that final answers are supported by coherent chains of evidence.
Across benchmarks covering open-web research, scientific reasoning, and financial analysis, MiroThinker-H1 achieves state-of-the-art performance on deep research tasks while maintaining strong results on specialized domains. We also release MiroThinker-1.7 and MiroThinker-1.7-mini as open-source models, providing competitive research-agent capabilities with significantly improved efficiency.

\vspace{2mm}


\coloremojicode{1F310} \textbf{Online Service}: \href{https://dr.miromind.ai/}{https://dr.miromind.ai}

\github{} \textbf{MiroThinker GitHub Repository}: \href{https://github.com/MiroMindAI/MiroThinker}{https://github.com/MiroMindAI/MiroThinker}

\github{} \textbf{MiroFlow GitHub Repository}: \href{https://github.com/MiroMindAI/MiroFlow}{https://github.com/MiroMindAI/MiroFlow}

\coloremojicode{1F917} \textbf{Model Weights}: \href{https://huggingface.co/miromind-ai/MiroThinker-1.7}{https://huggingface.co/miromind-ai/MiroThinker-1.7}

\end{abstract}

\maketitle
\vspace{3mm}
\begin{figure}[h]
\centering
\includegraphics[width=0.99\textwidth]{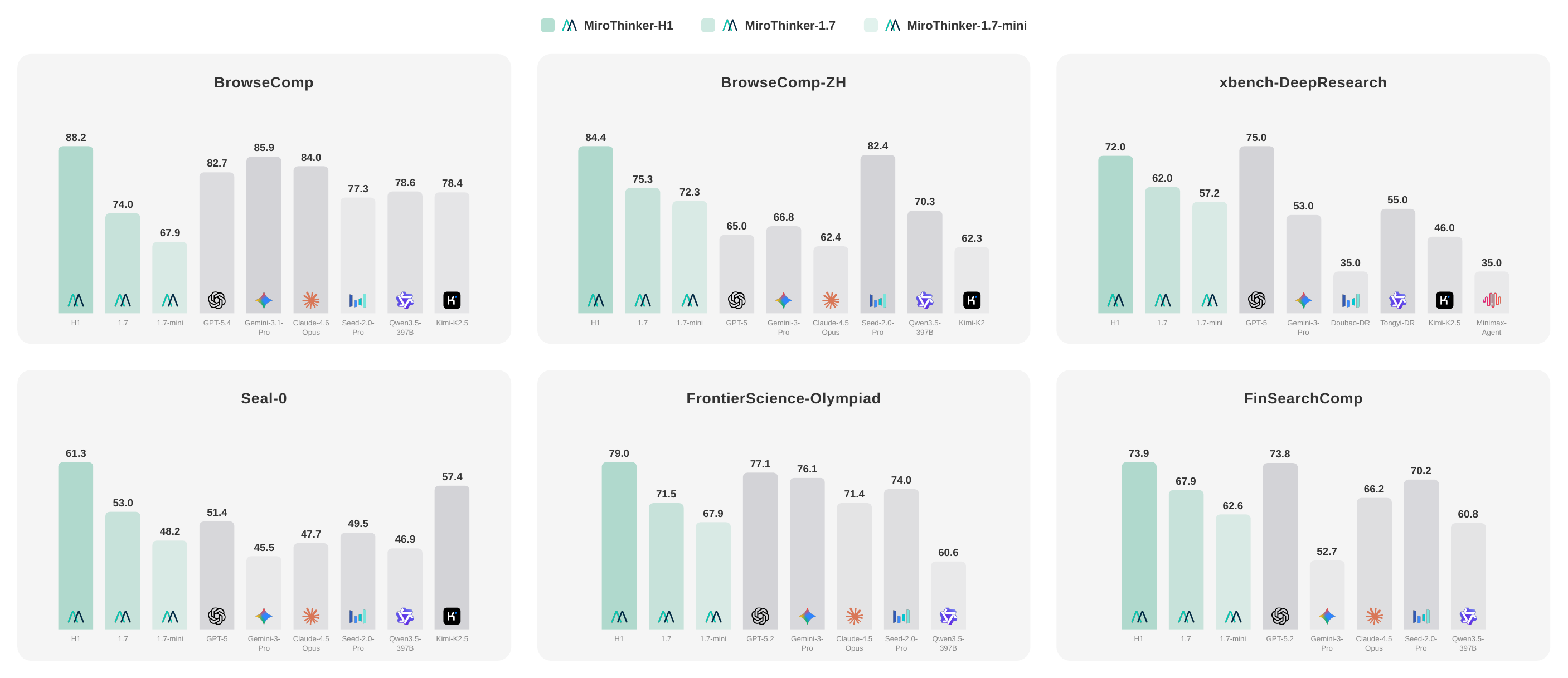}
\caption{Comparison of MiroThinker with state-of-the-art agents and agentic foundation models.}
\label{fig:teaser}
\end{figure}

\newpage
\section{Introduction}

Recent advances in large language models (LLMs) have significantly improved their ability to generate fluent text and answer a wide range of questions. However, many real-world problems, such as scientific analysis, financial reasoning, and open-ended research, require more than conversational ability~\citep{openai2026gpt54, minimax2026m25}. Solving these tasks typically involves long chains of reasoning, iterative information gathering, and the ability to verify intermediate conclusions before committing to a final answer.

These requirements have motivated the emergence of agentic AI systems, in which language models interact with tools, environments, and external knowledge sources to solve problems through multi-step reasoning and decision making~\citep{openai2026gpt54,  minimax2026m25, zeng2026glm5, qwenteam2026qwen35, bytedanceseed2026seed20pro, anthropic2026claude46opus, google2026gemini31pro}. While recent agent frameworks demonstrate promising capabilities, scaling the length of reasoning trajectories alone does not reliably improve performance. When intermediate steps are inaccurate or poorly grounded, longer interaction trajectories may instead accumulate noise, propagate errors, and ultimately degrade solution quality.

In this work, we argue that improving long-horizon reasoning requires scaling effective interaction rather than simply increasing interaction length. Effective interaction depends on two key factors: (1) strong atomic agentic capabilities at each step, including planning, reasoning, and effective tool execution; and (2) verifiable mechanisms that allow the system to verify and refine reasoning trajectories during problem solving. Without these elements, additional interaction steps may increase computational cost without meaningfully improving reasoning quality.
Motivated by this insight, we introduce MiroThinker-1.7, a deep research agent with stronger step-level reasoning capabilities, thereby enabling more effective interaction scaling. Building on this foundation, MiroThinker-H1 further introduces a verification-centric reasoning mode that enables more reliable long-horizon problem solving.

First, we develop a fully integrated training pipeline that connects multiple training stages, including mid-training, supervised fine-tuning, preference optimization, and reinforcement learning. In particular, we introduce an agentic mid-training stage designed to strengthen the model’s step-level agentic atomic capabilities, including planning, reasoning, tool use, and answer summarization. This stage leverages large-scale supervision emphasizing task decomposition, structured reasoning, and tool interaction patterns. By exposing the model to diverse forms of agentic supervision -- such as cold-start planning, context-conditioned reasoning, and intermediate summarization -- the model learns to make more reliable reasoning and action decisions at each step of the problem-solving process. As a result, each interaction step becomes more reliable and informative, which improves the scalability and effectiveness of interactive reasoning. Empirically, MiroThinker-1.7 demonstrates substantially stronger reasoning performance compared to MiroThinker-1.5, while requiring fewer reasoning turns to solve complex tasks.

Second, we introduce a heavy-duty reasoning mode that integrates verification into the reasoning process at both local and global levels, as shown in Figure~\ref{fig:training_framework}. At the local level, intermediate reasoning steps -- such as planning decisions, tool invocations, or hypothesis updates -- are evaluated and refined during inference, enabling the model to reconsider alternative actions and correct potential errors early in the reasoning trajectory. At the global level, the system audits the overall reasoning trajectory and compares candidate solution paths to ensure that the final answer is supported by the most coherent and well-grounded chain of evidence. Together, these mechanisms support more reliable long-horizon reasoning in complex real-world environments.

We evaluate the MiroThinker family across a diverse set of benchmarks covering open-web research, scientific reasoning, and financial analysis. As shown in Figure~\ref{fig:teaser}, our flagship system MiroThinker-H1 achieves 88.2 and 84.4 on BrowseComp~\citep{wei2025browsecomp} and BrowseComp-ZH~\citep{zhou2025browsecomp_zh}, respectively, outperforming best open-source and commercial research agents. In addition, the system demonstrates strong results on specialized benchmarks such as FrontierScience-Olympiad\citep{wang2025frontierscience} and FinSearchComp~\cite{hu2025finsearchcomp}, highlighting its ability to tackle complex reasoning tasks in scientific and financial domains. Our open-source models, MiroThinker-1.7 and MiroThinker-1.7-mini, remain highly competitive while maintaining significantly improved efficiency.
To sum, these results suggest that combining agent-native training with verification-centric reasoning provides a promising path toward building AI systems capable of sustained long-chain reasoning and reliable problem solving in complex real-world environments.

\section{Related Works}

\paragraph{Agentic Large Language Models} 

Recent advances in LLMs have increasingly focused on enabling \emph{agentic behavior}, 
where models autonomously decompose complex goals into sub-tasks, invoke external tools, 
and iteratively refine intermediate decisions based on environmental 
feedback~\citep{openai2025gpt5, anthropic2025claude4_5, liu2024deepseekv3, yang2025qwen3, team2025kimik2}. 
Unlike conventional chatbots that primarily rely on single-step responses to user inputs, agentic LLMs maintain persistent reasoning traces across multiple steps and dynamically coordinate tool execution.

Recent frontier models increasingly integrate such capabilities either during training 
or through tightly coupled inference frameworks. Representative examples include 
GPT-5.4~\citep{openai2026gpt54}, 
Claude-4.6~\citep{anthropic2026claude46opus}, 
Gemini-3.1 Pro~\citep{google2026gemini31pro}, 
DeepSeek-V3.2~\citep{liu2025deepseekv32}, 
Qwen3.5-397B~\citep{qwenteam2026qwen35}, 
GLM-5.0~\citep{zeng2026glm5}, 
Minimax-M2.5~\citep{minimax2026m25}, 
Seed-2.0-Pro~\citep{bytedanceseed2026seed20pro},  
and Kimi-K2.5~\citep{team2025kimik25}. 
These models demonstrate strong performance across reasoning, coding, and multimodal benchmarks, 
while supporting long-context processing and integrated tool execution.

Collectively, these developments indicate a paradigm shift in which foundation models are evolving from passive language generators into \emph{general-purpose autonomous agents} capable of executing complex workflows and interacting with real-world environments.

\paragraph{Deep Research Agents}  
Building on the emergence of agentic LLMs, recent work has introduced \emph{deep research agents}, a class of LLM-based systems designed for open-ended knowledge synthesis tasks requiring long-horizon reasoning and intensive information retrieval. 
Rather than answering questions solely from pre-trained knowledge, these systems actively acquire external information, iteratively refine hypotheses, and synthesize evidence from multiple sources to produce structured research outputs.

Industrial systems have begun deploying such capabilities at scale. Representative examples include OpenAI Deep Research~\citep{openai2025deepresearch}, Claude Research~\citep{anthropic2025claude_research}, Kimi-Researcher~\citep{moonshot2025kimi_researcher}, and Grok DeepSearch~\citep{xai2025grok3}, all of which couple LLMs with integrated web browsing and multi-step planning to support autonomous, end-to-end research workflows.
Meanwhile, the research community has explored a variety of approaches for building open deep research agents. Works such as MiroThinker~\cite{team2025mirothinker}, WebThinker~\citep{li2025webthinker}, Tongyi DeepResearch~\citep{team2025tongyi}, DeepResearcher~\citep{zheng2025deepresearcher} investigate different strategies for enabling long-horizon research workflows. Notably, agentic mid-training has emerged as a common strategy for enhancing model agent capabilities, as adopted by Tongyi DeepResearch~\citep{team2025tongyi,su2025scaling}, REDSearcher~\citep{chu2026redsearcher}, Step-DeepResearch~\citep{hu2025step}, among others.
These efforts highlight a broader shift toward LLM systems that function as autonomous research assistants, capable of long-horizon information gathering, reasoning, and synthesis for complex open-ended tasks.


\section{Agentic Workflow}
\label{sec:agentic_workflow}

Deep research tasks require acquiring, verifying, and synthesizing evidence from diverse external sources across many reasoning steps, a process that fundamentally cannot be reduced to a single forward pass through a language model.
MiroThinker-1.7 is designed around this principle, implementing an iterative agent–environment interaction loop in which the model alternates between reasoning, tool invocation, and observation until it has gathered sufficient evidence to produce a final answer.
This section describes the three components that make this possible: the formal interaction loop (\S\ref{sec:formulation}), the modular tool interface that connects the agent to the external world (\S\ref{sec:tool_interface}), and the implementation strategies that sustain long-horizon trajectories within a fixed token budget (\S\ref{sec:implementation_details}).

\subsection{Formulation}
\label{sec:formulation}

MiroThinker-1.7 builds on the ReAct paradigm~\citep{yao2022react}, extending it with context management and tool-call correction within a single-agent architecture.
The agent operates in a dual-loop structure: an outer \emph{episode loop} that handles trajectory-level restarts, and an inner \emph{step loop} that drives reasoning, tool invocation, and observation within each episode.

\paragraph{Step Loop}
Within episode $e$, at step $t$ the framework accumulates a trajectory log
\begin{equation}
  H_t^{(e)} = \bigl\{(T_1, A_1, O_1),\; \ldots,\; (T_{t-1}, A_{t-1}, O_{t-1})\bigr\},
\end{equation}
where $T_i$, $A_i$, and $O_i$ denote the thought, action, and observation at step $i$, respectively.
The trajectory log records all raw outputs; however, the agent does not reason over $H_t^{(e)}$ directly.
Instead, a context operator $\Phi_t$ transforms the log into an effective context $C_t^{(e)}$ that fits within the token budget while preserving essential information.

We define a sliding-window index set
\begin{equation}
  S_t(K) = \bigl\{\, i \in \{1, \ldots, t{-}1\} \mid i \ge t - K \,\bigr\},
\end{equation}
which selects the $K$ most recent steps.
The context operator applies truncation within the window and masking outside it:
\begin{equation}
  \Phi_t(O_i) =
  \begin{cases}
    \mathrm{Trunc}_L(O_i), & i \in S_t(K),     \\
    \varnothing,            & \text{otherwise},
  \end{cases}
\end{equation}
where $\mathrm{Trunc}_L(\cdot)$ clips an observation to at most $L$ tokens and $\varnothing$ denotes omission from the context window.
Note that when $t \le K$, we have $S_t(K) = \{1, \ldots, t{-}1\}$, so all observations are retained (subject only to truncation) during the early steps of a trajectory.

\begin{figure}[t]
\centering
\includegraphics[width=0.95\textwidth]{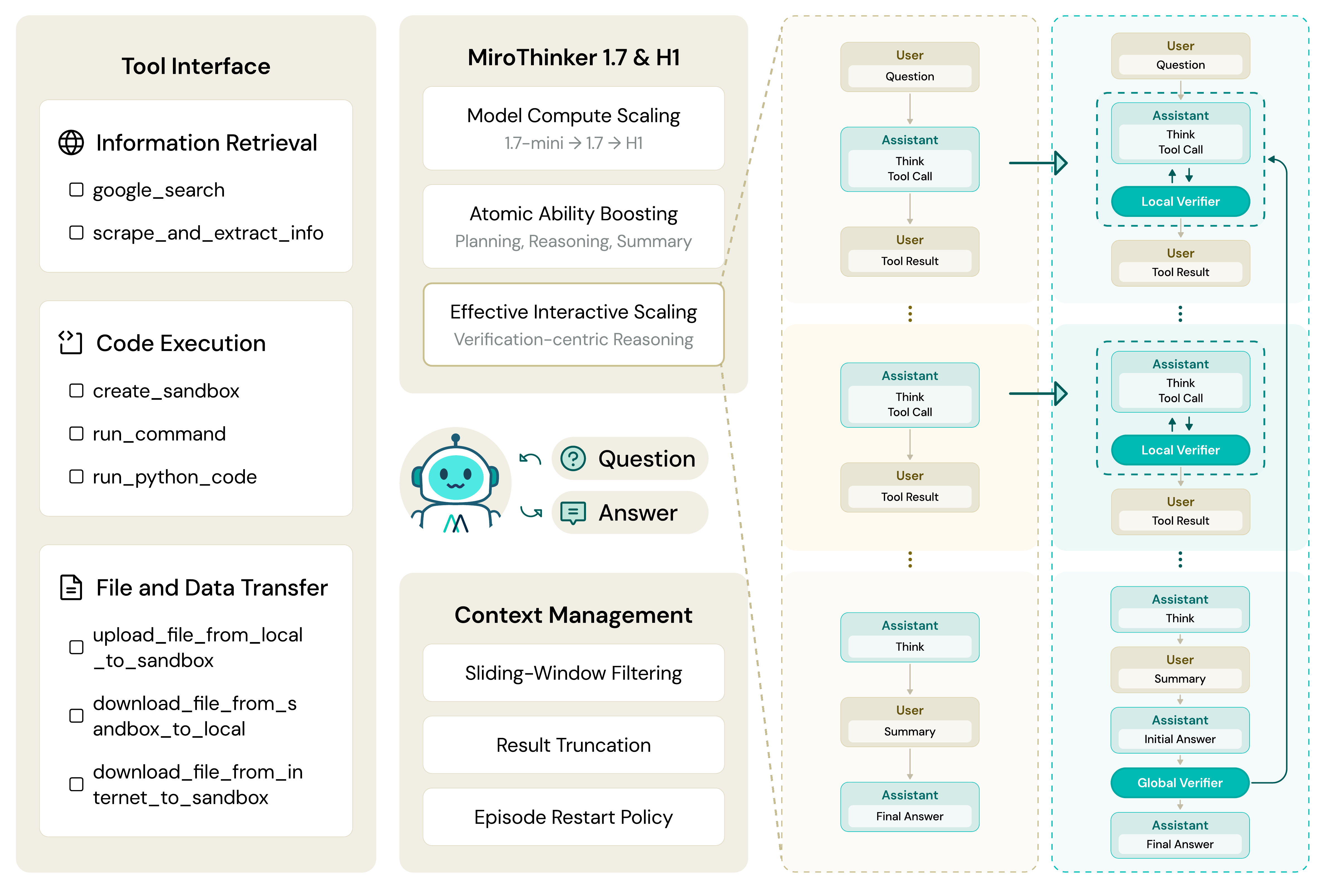}
\caption{The overview of MiroThinker-1.7 \& H1.}
\label{fig:training_framework}
\end{figure}

The effective context retains the complete thought-and-action trace while applying $\Phi_t$ only to observations:
\begin{equation}
  C_t^{(e)} = \bigl\{\bigl(T_i,\; A_i,\; \Phi_t(O_i)\bigr)\bigr\}_{i=1}^{t-1}.
\end{equation}

All reasoning and action selection operate on this managed view:
\begin{equation}
  T_t = f_{\theta}\!\bigl(q,\; C_t^{(e)}\bigr),
  \qquad
  A_t = \pi_{\theta}\!\bigl(C_t^{(e)},\; T_t\bigr).
\end{equation}
The environment executes the action and returns an observation $O_t = \mathrm{Tool}(A_t)$, after which the trajectory log is extended:
\begin{equation}
  H_{t+1}^{(e)} = H_t^{(e)} \cup \bigl\{(T_t,\, A_t,\, O_t)\bigr\}.
\end{equation}

\paragraph{Episode Loop}
The first episode is initialized with the query alone:
$C_0^{(1)} = \{q\}$.
If an episode exhausts its maximum turn budget $T_{\max}$ without producing a valid answer, whether due to reaching $T_{\max}$ or a persistent final-answer format error, the agent transitions to a new episode. We set the maximum number of new episode retries as a parameter $R_{\max}$.
The next episode re-initializes the agent with the original query alone:
\begin{equation}
  C_0^{(e)} = \bigl\{q\bigr\},
  \qquad e > 1,
\end{equation}
which is identical in form to the first-episode initialization, effectively discarding all information from the preceding trajectory.
This clean-slate restart avoids any bias from a potentially degraded context and ensures that the agent remains within the context budget.

This dual-loop design enables dynamic, evidence-grounded reasoning at a scale that would otherwise be unattainable.

\subsection{Tools}
\label{sec:tool_interface}

MiroThinker-1.7 implements its agent framework directly within the MiroThinker codebase, optimized for training data collection, model evaluation, and iterative development of our research agents.
Readers interested in a more general-purpose agent framework supporting richer agentic topologies may refer to MiroFlow~\citep{miromind2025miroflow}.
As shown in Figure~\ref{fig:training_framework}, tools are organized into three functional categories, each encapsulating a specific external capability.

\paragraph{Information Retrieval}
Open-web knowledge acquisition is supported by two tightly coupled tools.
The search tool (\texttt{google\_search}) submits structured queries to a Google-based backend and returns ranked results including titles, URLs, and snippets, giving the agent a broad view of relevant sources before selecting specific pages for full content extraction.

The scraping tool (\texttt{scrape\_and\_extract\_info}) then performs targeted content extraction from specified URLs. Retrieval proceeds through a multi-level fallback pipeline, with Jina serving as the primary scraping backend. Regardless of which backend successfully retrieves the page, the agent then passes the raw content to a lightweight language model that distills it into focused, task-relevant evidence as directed by the agent, avoiding the need to expose lengthy web documents directly to the model's context. 
This combination of layered retrieval robustness and LLM-mediated summarization allows the agent to reliably acquire and digest web content even when individual sources or backends are unavailable.

\paragraph{Code Execution}
An E2B Linux sandbox provides an isolated and reproducible runtime for command and code execution. The agent creates a sandbox instance via \texttt{create\_sandbox} and subsequently issues shell commands (\texttt{run\_command}) or executes Python scripts (\texttt{run\_python\_code}) within it, enabling safe interaction with system-level resources such as file I/O, numerical computation, and data processing.

\paragraph{File and Data Transfer}
Bidirectional file transfer utilities bridge the sandbox and the external world.
\texttt{upload\_file\_from\_local\_to\_\allowbreak sandbox} and \texttt{download\_file\_from\_\allowbreak sandbox\_to\_local} handle local transfers, while \texttt{download\_file\_from\_\allowbreak internet\_to\_\allowbreak sandbox} allows the agent to directly retrieve remote assets such as datasets or documents at inference time.

\subsection{Implementation Details}
\label{sec:implementation_details}

This section describes the practical choices underlying the agent's operation, including the context management strategies introduced in Section~\ref{sec:formulation} and additional robustness mechanisms.

\paragraph{Sliding-Window Filtering}
The sliding-window size is set to $K = 5$ (i.e., the five most recent observations) in all experiments. 
The key empirical insight is that the agent's decisions at step $t$ depend primarily on recent observations; retaining distant outputs yields diminishing returns at significant token cost.
Crucially, retaining the full thought-and-action trace means the agent preserves its global reasoning context and can refer back to earlier decisions, while concentrating its observation window on the most actionable recent evidence.
This strategy introduces negligible performance degradation while enabling substantially longer and deeper agentic trajectories.

\paragraph{Result Truncation}
Tools such as \texttt{run\_command} and \texttt{run\_python\_code} can produce outputs of unbounded length that risk exhausting the remaining context budget in a single step.
The truncation limit $L$ within the context operator $\Phi_t$ (Section~\ref{sec:formulation}) is applied per tool output to enforce a hard ceiling on individual observation size.
A \texttt{[Result truncated]} marker is appended whenever truncation occurs, signaling to the model that the output has been shortened so that it can issue a more targeted follow-up action if needed.

\paragraph{Episode Restart Policy}
The maximum turn budget $T_{\max}$ is set on a per-benchmark basis to accommodate varying task complexity.
When an episode reaches $T_{\max}$ without producing a final answer, the agent discards all prior state and restarts from the original query alone, as defined by the episode transition in Section~\ref{sec:formulation}.
On the final episode, the agent no longer defers answer generation: even if $T_{\max}$ is reached again, it attempts to produce an answer and falls back to the best intermediate answer extracted from the trajectory, ensuring the agent always returns its best available answer rather than failing silently.
This mechanism is particularly effective in conjunction with sliding-window filtering: long trajectories that exhaust $T_{\max}$ tend to accumulate stale context, and a clean restart allows the agent to re-engage the problem with a fresh context budget.

\paragraph{Tool Call Robustness}
In practice, language models occasionally produce malformed tool invocations, including incorrect server routing, hallucinated tool names, or mismatched parameter names.
We intercept and automatically correct such mistakes at the framework level before execution, improving reliability across long-horizon trajectories where accumulated failures would otherwise derail the agent's reasoning.

\paragraph{Benchmark Contamination Prevention}
We actively monitor for potential benchmark contamination and block access to identified sources at the infrastructure level. Known sources of leakage, such as HuggingFace dataset pages where benchmark questions and ground-truth answers are publicly hosted, are explicitly blocked. Whenever a new domain is found to expose benchmark content, it is immediately added to a blocklist that applies uniformly across all tools, ensuring that no retrieval pathway can circumvent the restriction during evaluation.

\begin{figure}[t]
\centering
\includegraphics[width=0.95\textwidth]{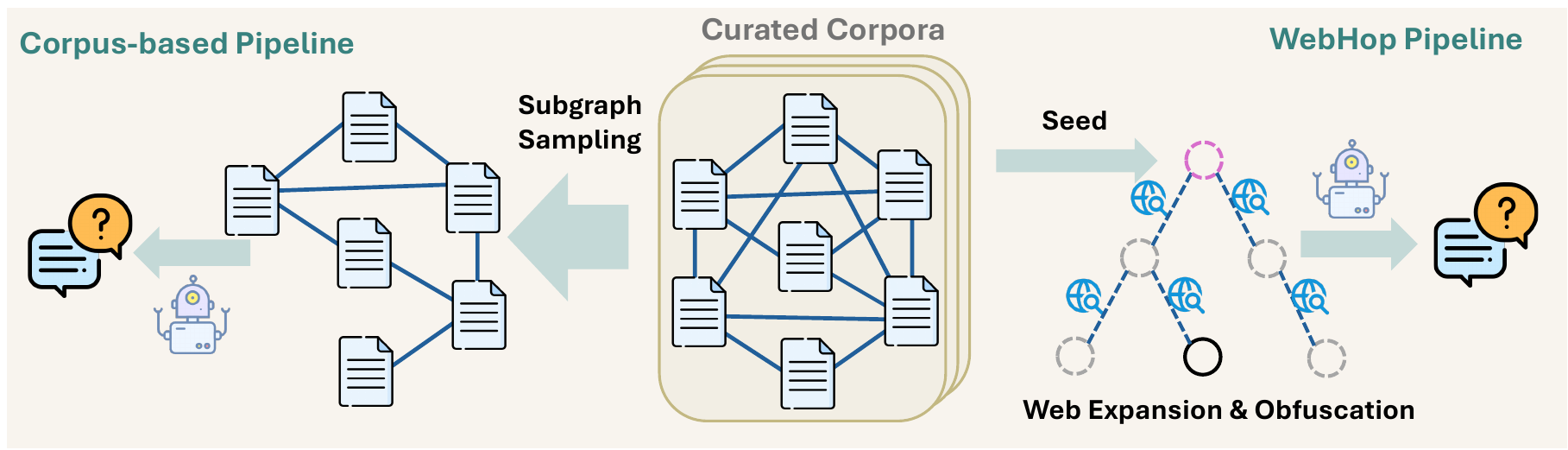}
\caption{Overview of the dual-pipeline QA synthesis framework. The Corpus-based Pipeline (left) focuses on topical breadth and high-throughput generation from document subgraphs. The WebHop Pipeline (right) constructs calibrated reasoning trees with web-augmented expansion and hierarchical verification to ensure reasoning rigour and controllable complexity.}
\label{fig:qa_pipeline}
\end{figure}

\section{High-Quality QA Construction} 
We design a QA synthesis framework with two complementary pipelines: a \textbf{Corpus-based Pipeline} for efficient large-scale generation from structured knowledge graphs, and a \textbf{Web-Augmented Multi-hop Pipeline} (WebHop) that combines web knowledge expansion with explicit difficulty control. The two pipelines jointly provide \emph{breadth} and \emph{depth}: the Corpus-based Pipeline produces high-volume QA pairs with \emph{diverse question structures and reasoning patterns} over curated corpora to build foundational reasoning capability, while WebHop generates fewer but precisely calibrated questions with verified multi-hop structure and open-web grounding. In training, Corpus-based output dominates early stages, and WebHop output is progressively introduced to push the model toward harder and more realistic challenges.

\subsection{Corpus-based Pipeline}
\label{sec:basic_pipeline}

Following MiroThinker 1.0~\cite{team2025mirothinker}, we construct document corpora from highly interlinked sources (\textit{e.g.}, Wikipedia, OpenAlex), preserving hyperlink topology. For each seed document, we sample a connected subgraph via internal hyperlinks, extract cross-document factual statements, and prompt a strong LLM to synthesize multi-hop QA pairs. This pipeline achieves high throughput and broad coverage, while inducing \emph{diverse question forms and reasoning patterns} via prompt-driven diversification and obfuscation; however, difficulty control remains implicit, there is no structural enforcement of reasoning depth or systematic control over information leakage.

\subsection{Web-Augmented Multi-hop Pipeline (WebHop)}
\label{sec:webhop_pipeline}

The WebHop Pipeline addresses these limitations through three mechanisms: structured reasoning graphs, web-based knowledge expansion, and hierarchical difficulty control.

\paragraph{Structured Multi-hop Graphs.}
We construct directed reasoning trees rooted at the answer entity, where each edge represents a verifiable semantic relationship. Tree depth controls the number of reasoning hops, and fact extraction is restricted to parent-child edges, preventing shortcut solutions that bypass the intended reasoning path.

\paragraph{Web-based Semantic Expansion.}
To broaden the knowledge distribution beyond curated corpora, we expand reasoning graphs via live web search. Root entities are drawn from existing knowledge bases to ensure verifiable answers; child nodes are then expanded by retrieving and selecting semantically related web pages, with encyclopedic sources excluded to introduce genuinely novel knowledge. This grounds QA pairs in diverse, real-world content that mirrors inference-time conditions.

\paragraph{Hierarchical Solvability Verification.}
We ensure each question is both solvable and non-trivial through verification at every level of the reasoning graph. For each parent-child relationship, we verify that knowing the children suffices to narrow the candidate set for the parent to a small range--concretely, a search agent given the child entities should locate the parent within bounded candidates. For the root entity, a stricter criterion applies: it must be uniquely identifiable from its first-hop neighbors alone, verified by prompting an LLM to infer the hidden root from an anonymized fact table. Failed samples are rejected before expensive downstream steps, maintaining both quality and efficiency.

\paragraph{Adaptive Leaf Obfuscation.}
Leaf entities most likely to leak the answer through surface associations (\textit{e.g.}, ``\textit{Louvre Pyramid}'' $\to$ ``\textit{Louvre Museum}'') are replaced with functional descriptions that expand the set of plausible referents (\textit{e.g.}, ``\textit{a royal residence in southern England}''). Each description is automatically verified: if an LLM can directly identify the original entity from the description, it is rejected and regenerated.

\paragraph{QA Generation.}
Given the verified and obfuscated reasoning graph, a strong LLM is used to generate multi-hop questions. The root entity serves as the answer, while leaf-level constraints enforce full-depth traversal of the graph. Additionally, only facts along the graph edges are allowed to be used in the question.

\subsection{Difficulty-Adaptive Filtering}
\label{sec:difficulty_filtering}

Beyond generation-time controls, we apply post-hoc filtering using search agents of varying capability. Questions that are solvable by weaker agents are allocated to earlier training stages (\textit{e.g.}, supervised fine-tuning), while those resisting stronger agents are reserved for later stages (\textit{e.g.}, reinforcement learning), producing a difficulty-graded corpus for curriculum-style training.

\section{Training Pipeline}

Based on the open-source Qwen3 MoE models~\cite{yang2025qwen3}, MiroThinker-1.7 is trained via a four-stage pipeline:
(1) Mid-training to strengthen atomic agentic capabilities, including planning, reasoning, tool call and answer summarization.
(2) Supervised fine-tuning to learn structured agentic interaction behaviors.
(3) Preference optimization to align the model’s decisions with task objectives and behavior preferences.
(4) Reinforcement learning to promote creative exploration and improve generalization in real-world environments.

\begin{figure}[t]
\centering
\includegraphics[width=0.99\textwidth]{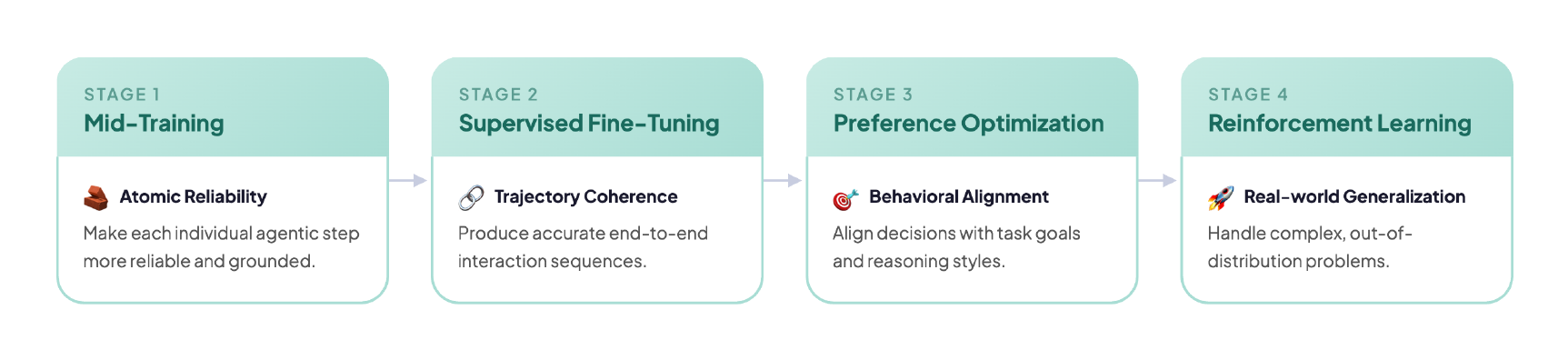}
\caption{The agentic training pipeline of MiroThinker-1.7.}
\label{fig:training_pipeline}
\end{figure}

\subsection{Agentic Mid-training}
The first-stage mid-training strengthens MiroThinker-1.7's agentic atomic capabilities, including planning, reasoning, tool use, and answer summarization. To achieve this, we scale up a large corpus of agentic supervision spanning single-turn \emph{planning}, \emph{reasoning} and \emph{summarization} data. These data target complementary aspects of agent behavior: cold-start planning from scratch, context-conditioned reasoning at intermediate steps of agent execution, and answer aggregation under limited or partial observations.  By exposing the model to these heterogeneous yet complementary forms of supervision, the mid-training stage equips MiroThinker-1.7 with stronger capabilities for structured problem solving, tool-aware reasoning, and coherent response generation in realistic agentic environments.

\paragraph{Agentic Planning Boosting.}
To build strong cold-start planning ability, we construct a large-scale single-turn planning corpus where the model learns to produce a structured plan and the first tool call given only the user query. The underlying data is drawn from diverse QA sources, including synthetic multi-hop QA and open-domain task data, and is deliberately diversified across domains to promote generalization. To ensure quality, we design a taxonomy-aware \emph{planner–judge} filtering pipeline. An LLM judge first classifies each problem into canonical categories (\textit{e.g.}, logic/mathematics, puzzle-style multi-hop retrieval, direct retrieval). We then apply category-specific criteria to reject common failure modes, such as verbatim query copying, over-constrained search formulations, premature entity guessing, and insufficient retrieval coverage. For knowledge-grounded planning, the judge further verifies whether the proposed plan can retrieve the core facts needed to solve the task. Rejected generations are re-sampled up to $K$ times. The data that still fail after $K$ attempts are discarded entirely, ensuring only high-quality plans enter the final corpus.

\paragraph{Agentic Reasoning and Summarization Sculpting.} Beyond cold-start planning, we train the model on \emph{interleaved reasoning and summarization} data constructed from multi-turn agent trajectories. Instead of supervising entire trajectories end-to-end, we isolate a single turn at step $k$ and rewrite it into a higher-quality target, conditioned on the full preceding context including dialogue history, prior tool calls, and intermediate outputs. Depending on the role of the selected turn, the rewrite targets either step-wise reasoning (\textit{e.g.}, evidence consolidation, tool-use decision making) or intermediate summarization (\textit{e.g.}, aggregating partial observations into a coherent answer). To improve generalization, we randomly apply context summarization strategies, so the model learns to reason and summarize flexibly under varying context conditions rather than relying on complete, well-structured trajectories.  Supervision is applied only to this rewritten turn, enabling the model to learn both skills under partially observed, dynamically evolving agent states without the noise inherent in full-trajectory training. To ensure quality, we source exclusively from successful trajectories with verified solution paths, and apply multi-level filtering that removes noisy or strategically inconsistent generations.

\paragraph{Training Objective.}
We train the model on the above agentic atomic data under a unified mid-training objective. In both settings, supervision is applied via next-token prediction over a single target assistant turn at step $k$, conditioned on the preceding context $C_{<k}$ (comprising the task instruction, prior reasoning, tool calls, and tool observations). For single-turn planning examples, $k{=}1$ and $C_{<1}$ reduces to the user query alone; for interleaved reasoning and summarization examples, $k > 1$ and $C_{<k}$ contains the full trajectory prefix up to that step. Formally, the mid-training objective is:
\begin{equation}
\mathcal{L}_{\text{mid}}(\theta) = -\mathbb{E}_{(C_{<k},\, y_k) \,\sim\, \mathcal{D}_{\text{mid}}} \left[\log \pi_\theta\!\left(y_k \mid C_{<k}\right)\right],
\end{equation}
where $y_k$ denotes the target assistant output at step $k$, \textit{e.g.}, a structured plan with the first tool call when $k{=}1$, or a rewritten reasoning/summarization turn when $k > 1$. Alongside the agentic atomic data, we also mix in general-purpose instruction-following and knowledge-intensive data to preserve the model's general capabilities and mitigate catastrophic forgetting. Together, these mid-training signals strengthen MiroThinker-1.7's agentic atomic capabilities and expand its domain coverage, making each individual step in the interaction more reliable and grounded. This establishes a stronger foundation for effective interactive scaling in subsequent post-training stages.


\subsection{Agentic Supervised Fine-tuning}
In the second stage, we apply supervised fine-tuning (SFT) to equip MiroThinker with structured agentic capabilities.
Specifically, the model is trained to replicate expert trajectories that require multi-step reasoning and tool interaction.

\paragraph{Data Construction}
We curate a large-scale SFT dataset $\mathcal{D}_{\text{SFT}} = \{(x_i, H_i)\}_{i=1}^N$, where each sample consists of a task instruction $x_i$ paired with an expert trajectory 
$H_i = \{(T_{i,t}, A_{i,t}, O_{i,t})\}_{t=1}^{T_i}$, represented as a sequence of thought--action--observation triplets.
We find that the raw trajectories, even when generated by strong LLMs, frequently contain considerable noise, including repetitive content within and across responses, malformed tool invocations (\emph{e.g.}, incorrect tool names or unparseable arguments), and undesirable behavioral patterns (\emph{e.g.}, invoking undefined tools or failing to retry after errors).
To address these issues, we apply a comprehensive rule-based filtering and data-cleaning pipeline to ensure the quality and consistency of the resulting SFT corpus.

\paragraph{Training Objective}
Each trajectory is formatted as a multi-turn conversation between a \emph{user} and an \emph{assistant}.
The user provides the initial task instruction $x$ along with the tool observations $O_t$ at each step, while the assistant generates the corresponding reasoning thoughts $T_t$ and tool calls $A_t$.
Note that tool execution is not performed during training; instead, the observations are pre-collected and provided as part of the input context.
Given $(x, H) \sim \mathcal{D}_{\text{SFT}}$, the training objective is to maximize the likelihood of the expert's thought and action sequences:
\begin{equation}
\mathcal{L}_{\text{SFT}}(\theta)
= -\mathbb{E}_{(x,H)} \left[ \sum_{t=1}^{T_H} \log \pi_\theta(T_t, A_t \mid x, H_{<t}) \right].
\end{equation}
This formulation casts the agent's imitation learning as standard dialogue-style SFT,
where tool outputs serve as user turns and the assistant is trained to produce the next next reasoning and tool call accordingly.

\subsection{Agentic Preference Optimization}
In the third stage, we further improve the model's decision-making ability through Direct Preference Optimization (DPO)~\citep{rafael2023dpo}, using preference data collected from the SFT model.

\paragraph{Data Collection}
We build a pairwise preference dataset
\begin{equation}
\mathcal{D}_{\text{PO}} = \{(x_i, H_i^{+}, H_i^{-})\}_{i=1}^{M} ,
\end{equation}
where each task instruction $x_i$ is paired with a preferred trajectory $H_i^{+}$ and a dispreferred trajectory $H_i^{-}$.
Each trajectory corresponds to a complete multi-step interaction consisting of thought, action, and observation. We determine preferences according to the following criteria:

(1) \textbf{Correctness-Based Ranking Without Structural Constraints.} We assign preferences primarily based on whether the final answer is correct. Some prior work relies on handcrafted heuristics or enforces fixed agentic patterns (\emph{e.g.}, predetermined planning length, step counts, or reasoning templates) to define preferences. However, we observe that such constraints can introduce systematic biases and limit generalization across different tasks and domains. We therefore do not impose any rigid structural requirements and instead use answer correctness as the sole ranking signal.

(2) \textbf{Quality Filtering for Trace Completeness.} We apply strict filtering to ensure the quality of both chosen and rejected trajectories. Specifically, a chosen trajectory must contain coherent reasoning, an explicit planning process, and a correct final answer. A rejected trajectory must also produce a valid final answer. Beyond these requirements, we further remove trajectories that exhibit surface-level issues such as repetition, truncation, or malformed output, so that only well-formed trajectories are kept in the dataset.

\paragraph{Training Objective}
We optimize the SFT model using DPO combined with an auxiliary SFT loss on preferred trajectories~\cite{liu2024rpo, wang2024mpo} to improve training stability and maintain behavioral consistency. Given a task instruction $x$ and a preference pair $(H^+, H^-)$, the DPO loss encourages the model to assign higher likelihood to the preferred trajectory relative to the reference model:
\begin{equation}
\mathcal{L}_{\text{DPO}}(x,H^+,H^-)
= - \log \sigma \!\left(
\beta \!\left[
(\log \pi_\theta(H^+|x) - \log \pi_\theta(H^-|x))
- (\log \pi_{\text{ref}}(H^+|x) - \log \pi_{\text{ref}}(H^-|x))
\right]
\right),
\end{equation}
where $\pi_{\text{ref}}$ is the frozen reference model and $\beta$ controls the degree of deviation from it.
The overall training objective combines the DPO loss with the SFT loss on preferred samples:
\begin{equation}
\mathcal{L}_{\text{PO}}(\theta)
= \mathbb{E}_{(x,H^+,H^-)} [\mathcal{L}_{\text{DPO}}(x,H^+,H^-)]
+ \lambda \, \mathcal{L}_{\text{SFT}}^{(+)}(\theta),
\end{equation}
where $\mathcal{L}_{\text{SFT}}^{(+)}$ is the SFT loss computed on preferred trajectories and $\lambda$ is the weighting coefficient.

\paragraph{Preference Distillation}
For MiroThinker-1.7-mini, we adopt a preference distillation strategy to transfer alignment signals from a stronger model during preference optimization. This design allows the policy to be guided not only by the preference signal from chosen–rejected pairs, but also by additional preference guidance derived from a more capable model. In practice, this encourages the MiroThinker-1.7-mini model to better align with the preference tendencies of the strong model while still learning from the preference data, leading to improved performance compared to standard DPO training.

\subsection{Agentic Reinforcement Learning}

\begin{figure}[t]
  \centering
  \begin{subfigure}{0.45\linewidth}
    \includegraphics[width=\linewidth]{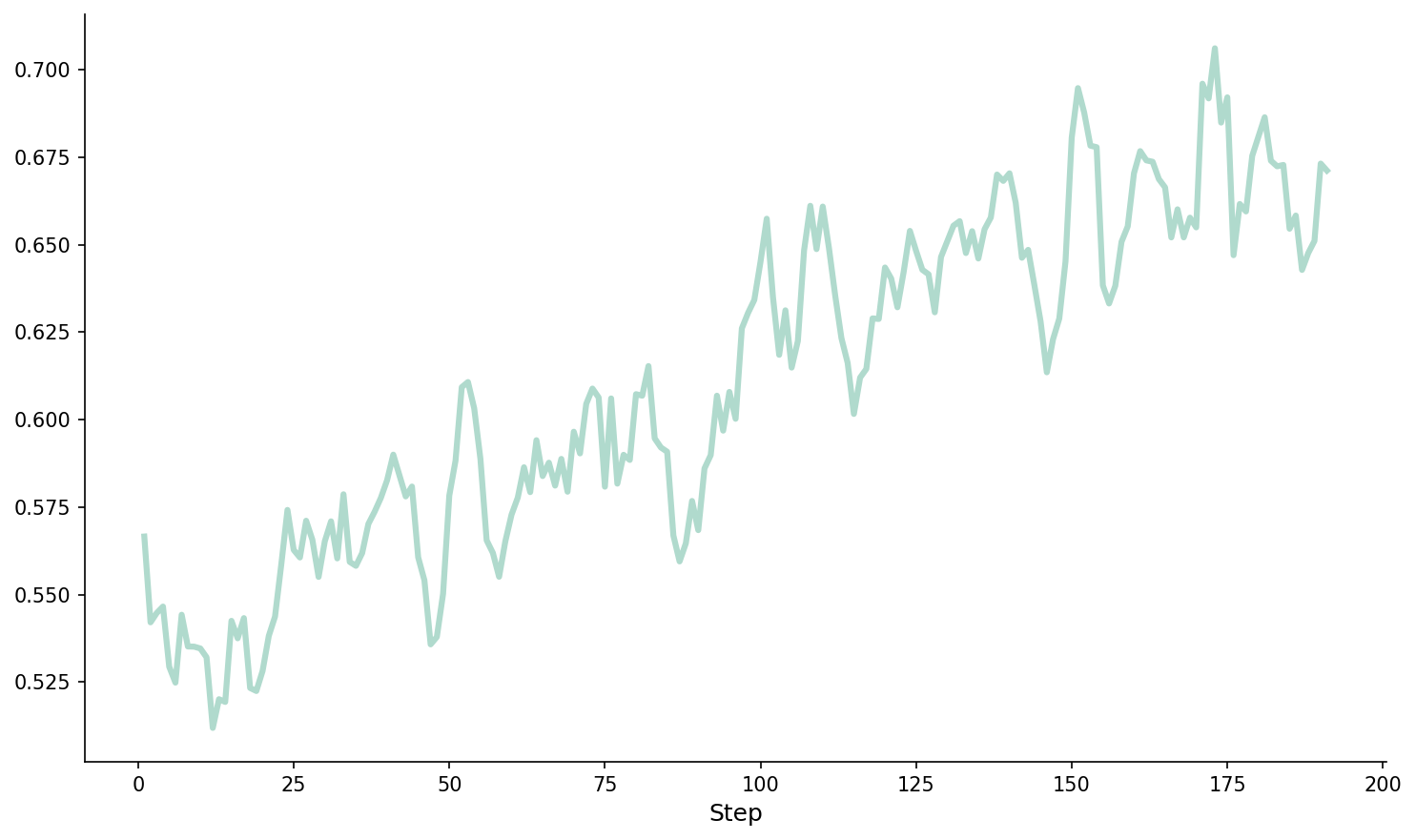}
    \caption{Training reward across training steps.}
    \label{fig:grpo_reward}
  \end{subfigure}
  \hspace{2em}
  \begin{subfigure}{0.45\linewidth}
    \includegraphics[width=\linewidth]{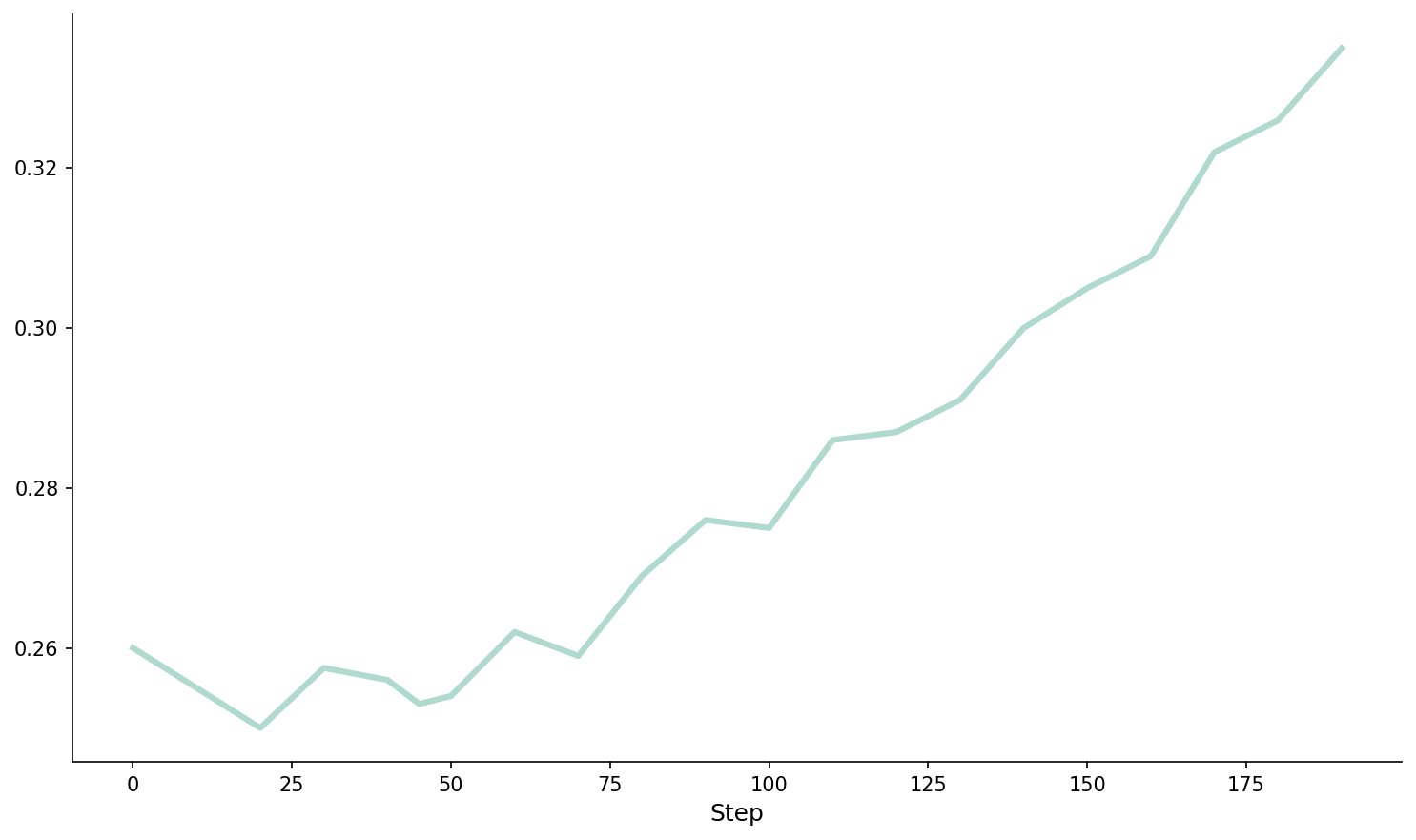}
    \caption{Val acc on BrowseComp-200 across training.}
    \label{fig:grpo_val}
  \end{subfigure}

\label{fig:grpo}
\caption{Training dynamics of MiroThinker-1.7-mini for GRPO Agentic RL.
BrowseComp-200 is our selected challenging subset from BrowseComp for faster evaluation during training. The plotted curves represent a running average with a window size of 5 to highlight the optimization trends.
}

\end{figure}

In the final phase of training, we move beyond supervised objectives and instead allow the model to autonomously refine its behavior via trial-and-error within live environments.
This is achieved through RL, specifically Group Relative Policy Optimization (GRPO)~\cite{shao2024deepseekmath}, operated in a purely online fashion where each batch of collected rollouts is consumed for one single policy gradient step.

\paragraph{Infrastructure for Parallel Execution} 
central requirement for agentic RL at scale is the ability to run a large number of agent sessions simultaneously.
To this end, we engineer a distributed infrastructure spanning multi-source web retrieval, page-level content extraction and summarization.
Complementing these environments, we deploy a dedicated LLM-based answer verification module that adjudicates whether a noisy agent response matches the reference solution, operating under tight latency constraints to avoid becoming a training bottleneck.

\paragraph{Streaming Rollout Acceleration with Priority Scheduling}
MiroThinker~1.0 introduced streaming rollout acceleration, where workers pull tasks from a shared queue on a first-available basis and deposit completed trajectories into a buffer that triggers training once full.
Building upon this mechanism, we further introduce a priority scheduling strategy that promotes long-tailed rollouts so they are completed and incorporated into training as early as possible, preventing prolonged exclusion of difficult samples from distorting the training distribution.

\paragraph{Entropy Control}
Maintaining policy entropy is crucial for training stability. To mitigate premature entropy collapse, we introduce a targeted entropy control mechanism that applies an auxiliary KL penalty to tokens with low log probabilities, specifically within negative rollouts. This regularization prevents the model from continuously driving down the likelihood of these tokens, thereby sustaining a healthy level of exploration and stabilizing the overall optimization dynamics.

\paragraph{Reward Design and Training Objective}
We optimize our policy using GRPO coupled with a targeted entropy control mechanism. For a given question $x$, our reward function $R(x, H) = \alpha_c R_\text{correct}(H) - \alpha_f R_\text{format}(H)$ balances the persistent exploration of new solutions with instruction-following capabilities. GRPO samples a group of $G$ trajectories $\{H_1, \ldots, H_G\}$ per prompt and computes advantages relative to the group mean: $\hat{A}_i = R(x, H_i) - \frac{1}{G}\sum_{j=1}^G R(x, H_j)$. To maintain training stability and prevent premature entropy collapse, we integrate our entropy control directly into the token-level Kullback-Leibler (KL) regularization. The final objective is formulated as:
\begin{equation}
\mathcal{L}_\text{GRPO}(\theta) = \mathbb{E}_{x \sim \mathcal{D}} \mathbb{E}_{H \sim \pi_\theta} \left[ \hat{A}(x, H) \log \pi_\theta(H \mid x) - \sum_{t=1}^{|H|} \beta_\text{KL}(t, H) D_\text{KL}\big(\pi_\theta(\cdot \mid s_t) \,\|\, \pi_\text{ref}(\cdot \mid s_t)\big) \right],
\end{equation}
where $s_t$ denotes the context at step $t$. To penalize the continuous degradation of token likelihoods in unsuccessful trajectories, the dynamic penalty coefficient $\beta_\text{KL}(t, H) = \beta_0 + \beta_\text{ent} \mathbb{I}\big(\hat{A}(x, H) < 0 \land \log \pi_\theta(a_t \mid s_t) < \tau\big)$ applies an auxiliary KL penalty specifically to low-probability tokens ($\log \pi_\theta < \tau$) within negative rollouts ($\hat{A} < 0$).
\section{Heavy-duty Reasoning Mode}


In this section, we introduce a novel verification-centric reasoning scheme, our first systematic exploration of integrating explicit verification into long-horizon reasoning. As illustrated on the right side of Figure~\ref{fig:training_framework}, this reasoning mode is instantiated in MiroThinker-1.7 to produce MiroThinker-H1, which incorporates two new features for heavy-duty reasoning: Local Verifier and Global Verifier, which independently audit the step-level and the complete reasoning process, respectively.



\textbf{Local verification.} Under the standard ReAct paradigm, an agent naturally follows the highest-probability path suggested by the model. On hard problems, this probability bias can steer the agent into habitual thinking patterns. 
Local verification counters this by prompting the agent to explore more thoroughly and to selectively gather feedback from the environment.
This encourages a more thorough search of the solution space, rather than making exploration degenerate into repeated confirmation of the model’s own preferences.

\textbf{Global verification.} A long-underutilized fact is that verification is often easier than generation. Leveraging this generation-verification asymmetry, we introduce global verification, which organizes the full chain of evidence collected. If the evidence is insufficient, the system asks the agent to resample or complete its reasoning chain rather than deliver a premature answer. Under a controllable compute budget, the system ultimately selects the answer backed by the most complete and reliable evidence.

\begin{table}[t]
\centering
\footnotesize
\definecolor{rowhl}{RGB}{225,237,237}
\caption{Performance comparison across various agentic benchmarks. We report the latest publicly available benchmark results for competing models, with the corresponding scores taken from the  technical reports or model cards of other organizations. To minimize the impact of randomness from agent-environment interactions on benchmark performance evaluation, we report the average performance of our MiroThinker-1.7 models on each benchmark. We use avg@3 for BrowseComp, BrowseComp-ZH, Humanity’s Last Exam, and DeepSearchQA, and avg@8 for GAIA, xbench-DeepSearch-2510, and SEAL-0.}
\label{tab:benchmarks-fast}
\renewcommand\arraystretch{1.1}
\setlength{\tabcolsep}{4pt}
\begin{tabularx}{\linewidth}{lccccccc}
\toprule
\textbf{Benchmarks} &
\makecell{\textbf{Browse}\\\textbf{Comp}} &
\makecell{\textbf{Browse}\\\textbf{Comp-ZH}} &
\makecell{\textbf{Humanity's}\\\textbf{Last Exam}} &
\textbf{GAIA} &
\makecell{\textbf{xbench-DeepSearch}\\\textbf{-2510}} &
\textbf{SEAL-0} &
\textbf{DeepSearchQA} \\
\midrule
Qwen3.5-397B~\cite{qwenteam2026qwen35}      & 78.6 & 70.3 & 48.3 & -- & -- & 46.9 & -- \\
Tongyi-DeepResearch-30B~\cite{team2025tongyi}      & 43.4 & 46.7 & 32.9 & 70.9 & 55.0 & -- & -- \\
GLM-5.0~\cite{zeng2026glm5}                       & 75.9 & 72.7 & 50.4 & -- & -- & -- & -- \\
Minimax-M2.5~\cite{minimax2026m25}                 & 76.3 & -- & -- & -- & -- & -- & -- \\
DeepSeek-V3.2~\cite{liu2025deepseekv32}            & 67.6 & 65.0 & 40.8 & -- & -- & 49.5 & 60.9 \\
Kimi-K2.5~\cite{team2025kimik25}                   & 78.4 & -- & 50.2 & -- & 46.0 & 57.4 & 77.1 \\
Seed-2.0-Pro~\cite{bytedanceseed2026seed20pro}     & 77.3 & 82.4 & \textbf{54.2} & -- & -- & 49.5 & 77.4 \\
OpenAI-GPT-5~\cite{openai2025gpt5}           & 54.9 & 65.0 & 35.2 & 76.4 & \textbf{75.0} & 51.4 & 79.0 \\
OpenAI-GPT-5.4~\cite{openai2026gpt54}              & 82.7 & -- & 52.1 & -- & -- & -- & -- \\
Gemini-3.0-Pro~\cite{google2025gemini3pro}           & 59.2 & 66.8 & 46.9 & -- & 53.0 & 45.5 & 76.9 \\
Gemini-3.1-Pro~\cite{google2026gemini31pro}              & 85.9 & -- & 51.4 & -- & -- & -- & -- \\
Claude-4.5-Opus~\cite{anthropic2025claude45opus}     & 67.8 & 62.4 & 43.2 & -- & -- & 47.7 & 80.0 \\
Claude-4.6-Opus~\cite{anthropic2026claude46opus}     & 84.0 & -- & 53.1 & -- & -- & -- & \textbf{91.3} \\

\midrule
\rowcolor{rowhl} MiroThinker-1.7-mini              & 67.9 & 72.3 & 36.4 & 80.3 & 57.2 & 48.2 & 67.9 \\
\rowcolor{rowhl} MiroThinker-1.7              & 74.0 & 75.3 & 42.9 & {82.7} & 62.0 & 53.0 & 72.1 \\
\rowcolor{rowhl} MiroThinker-H1              & \textbf{88.2} & \textbf{84.4} & 47.7 & \textbf{88.5} & 72.0 & \textbf{61.3} & 80.6 \\ 

\bottomrule
\end{tabularx}
\end{table}

\begin{table}[t]
\centering
\footnotesize
\definecolor{rowhl}{RGB}{225,237,237}
\caption{Performance comparison across multiple professional-domain benchmarks, including scientific, financial, and medical domains. Some scores are taken from technical reports released by other organizations, while the results for Qwen3.5-397B are obtained through our internal evaluation. To mitigate the randomness introduced by agent–environment interactions during evaluation, we report the mean performance of our MiroThinker-1.7 models on each benchmark. We use avg@3 for FinSearchComp MedBrowseComp, and avg@8 for FrontierSci-Olympiad and SUPERChem.}
\label{tab:benchmarks-domain}
\renewcommand\arraystretch{1.1}
\setlength{\tabcolsep}{4pt}
\begin{tabular}{lcccc}
\toprule
\textbf{Model} &
\makecell{\textbf{FrontierSci}\\\textbf{-Olympiad}} &
\makecell{\textbf{SUPERChem}\\\textbf{(text only)}} &
\makecell{\textbf{FinSearchComp}\\\textbf{(T2/T3)}} &
\makecell{\textbf{MedBrowse}\\\textbf{Comp}} \\
\midrule
Qwen3.5-397B                & 60.6 & 49.6 & 60.8 & 47.9 \\
Seed-2.0-Pro                & 74.0 & 53.0 & 70.2 & --   \\
GPT-5.2-high                & 77.1 & 58.0 & 73.8 & --   \\
Claude-4.5-Opus             & 71.4 & 43.2 & 66.2 & --   \\
Gemini-3-Pro                & 76.1 & \textbf{63.2} & 52.7 & --   \\
Kimi-K2.5                   & --   & --   & 67.8 & --   \\
\midrule
\rowcolor{rowhl} MiroThinker-1.7-mini    & 67.9 & 36.8 & 62.6 & 48.2 \\
\rowcolor{rowhl} MiroThinker-1.7        & 71.5 & 42.1 & 67.9 & 54.2 \\
\rowcolor{rowhl} MiroThinker-H1    & \textbf{79.0} & 51.3 & \textbf{73.9} & \textbf{56.5} \\
\bottomrule
\end{tabular}
\end{table}

\section{Experiments}

\subsection{Experimental Setup}
\paragraph{Evaluation Benchmarks.}
We initialize our models from the Qwen3 MoE checkpoints~\cite{yang2025qwen3}.
We assess the resulting MiroThinker models on two categories of benchmarks.
The first category consists of \textbf{agentic benchmarks} that evaluate multi-step web browsing, information retrieval, and reasoning capabilities:
Humanity's Last Exam (HLE) \citep{phan2025hle},
BrowseComp \citep{wei2025browsecomp} and BrowseComp-ZH \citep{zhou2025browsecomp_zh},
GAIA \citep{mialon2023gaia},
DeepSearchQA \citep{gupta2025deepsearchqa},
WebWalkerQA \citep{wu2025webwalker},
FRAMES \citep{krishna2025fact},
and SEAL-0 \citep{pham2025sealqa}.
The second category consists of \textbf{domain-specific benchmarks} that assess expert-level reasoning in specialized fields:
FrontierSci-Olympiad \citep{wang2025frontierscience} for scientific reasoning,
SUPERChem \citep{zhao2025superchem} for chemistry,
FinSearchComp \citep{hu2025finsearchcomp} for finance,
and MedBrowseComp \citep{chen2025medbrowsecomp} for medicine.
Following standard evaluation protocols for consistency with prior work, we report results on the 2,158 text-only subset of Humanity's Last Exam, the text-only subset of SUPERChem, and the T2/T3 subset of FinSearchComp. For all other benchmarks, results are reported on the complete test set.
To mitigate potential data contamination (\emph{e.g.}, retrieving benchmark answers from HuggingFace), we explicitly block access to the relevant website within the tool environment.

\paragraph{Evaluation Protocol.}
All benchmark results are obtained using a straightforward ReAct-style agent, which allows us to directly reflect the capability of our MiroThinkers.
We use fixed inference hyperparameters throughout to ensure stability and reproducibility: temperature = 1.0, top-p = 0.95, context length = 256K tokens, and maximum output length = 16,384 tokens.
The maximum number of interaction turns $T_{max}$ is set to 200 for most benchmarks, except for BrowseComp, BrowseComp-ZH, and DeepSearchQA where it is set to 300. We set the maximum number of new episode retries $R_{max}=5$ and the context management retention budget $K=5$.
For benchmarks with high per-question variance, we perform $k$ independent trials and report the mean score, denoted as avg@k. The specific $k$ values for all benchmarks are detailed in the caption of Table~\ref{tab:benchmarks-fast} and ~\ref{tab:benchmarks-domain}.
All benchmark performances are evaluated using an LLM-as-a-Judge approach.
Specifically, GAIA, WebWalkerQA, DeepSearchQA, BrowseComp, and BrowseComp-ZH are judged by gpt-4.1-2025-04-14, while Humanity's Last Exam follows its official protocol using o3-mini-2025-01-31.



\subsection{Overall Performance}

MiroThinker-H1 achieves 88.2 on BrowseComp~\cite{wei2025browsecomp} and 84.4 on BrowseComp-ZH~\cite{zhou2025browsecomp_zh}, outperforming strong commercial agents including Gemini-3.1-Pro~\cite{google2026gemini31pro} (85.9) and Claude-4.6-Opus~\cite{anthropic2026claude46opus} (84.0) on BrowseComp, and Seed-2.0-Pro~\cite{bytedanceseed2026seed20pro} (82.4) on BrowseComp-ZH.
MiroThinker-H1 also establishes a new state-of-the-art on the GAIA benchmark~\cite{mialon2023gaia} with a score of 88.5, surpassing the previous leading model, OpenAI-GPT-5 (76.4), by 12.1 percentage points.
On xbench-DeepSearch~\cite{chen2025xbench}, MiroThinker-H1 scores 72.0, narrowing the gap with OpenAI-GPT-5 (75.0).
Furthermore, MiroThinker-H1 achieves 61.3 on SEAL-0~\cite{pham2025sealqa}, setting a new best result among all evaluated models, and scores 80.6 on DeepSearchQA~\cite{gupta2025deepsearchqa}. 
Notably, MiroThinker-1.7-mini, with only 3B activated parameters, achieves competitive results across all benchmarks, outperforming strong models such as GPT-5~\cite{openai2025gpt5} and DeepSeek-V3.2~\cite{liu2025deepseekv32} on BrowseComp-ZH and GAIA. MiroThinker-1.7 further narrows the gap with the best proprietary systems across the board.

\subsection{Professional-domain Performance} 
We further evaluate MiroThinker on a set of challenging professional-domain benchmarks spanning scientific, chemical, financial, and medical tasks. As shown in Table~\ref{tab:benchmarks-domain}, these benchmarks include FrontierSci-Olympiad (scientific reasoning), SUPERChem (chemistry reasoning), FinSearchComp (financial search and analysis), and MedBrowseComp (medical browsing and synthesis). 
Overall, the MiroThinker series demonstrates strong performance across these specialized domains. In particular, MiroThinker-H1 achieves the best results on three out of four benchmarks, including FrontierSci-Olympiad (79.0), FinSearchComp (73.9), and MedBrowseComp (56.5). 

Notably, on FrontierSci-Olympiad, MiroThinker-H1 surpasses strong frontier models such as GPT-5.2-high (77.1) and Gemini-3-Pro (76.1), highlighting strong capabilities in complex scientific reasoning.
MiroThinker also maintains competitive performance across the other professional-domain benchmarks. MiroThinker-H1 achieves the highest scores on FinSearchComp and MedBrowseComp among the compared models, while remaining competitive on SUPERChem where Gemini-3-Pro obtains the top result. Taken together, these results demonstrate that MiroThinker performs robustly across multiple specialized domains, highlighting its effectiveness on knowledge-intensive tasks in professional domains.


\subsection{Long Report Evaluation}

\begin{table}[t]
\centering
\footnotesize
\definecolor{rowhl}{RGB}{225,237,237}
\caption{Long report evaluation on 50 deep research queries automatically generated using the DeepResearchEval query generation framework.}
\label{tab:model-comparison-report}
\renewcommand\arraystretch{1.1}
\setlength{\tabcolsep}{4pt}
\begin{tabular}{lccc}
\toprule
\textbf{Model} & \textbf{Report} & \textbf{Factuality} & \textbf{Overall} \\
\midrule

Grok Deep Research & 57.4 & 58.0 & 57.7 \\
Manus-1.6-Max Wide Research & 53.6 & 76.4 & 65.0 \\
Doubao Deep Research & 65.8 & 65.8 & 65.8 \\
Qwen-3.5-Plus Deep Research & 62.4 & 73.6 & 68.0 \\
Claude-Opus-4.6 Research & 69.9 & 66.2 & 68.0 \\
MiniMax-M2.5 Research & 62.2 & 76.4 & 69.3 \\
GLM-5 Agent & 66.0 & 72.7 & 69.4 \\
Kimi-K2.5 Deep Research & 76.0 & 64.1 & 70.0 \\
Gemini-3.1-Pro Deep Research & 72.3 & 73.3 & 72.8 \\
ChatGPT-5.4 Deep Research & 76.4 & \textbf{85.5} & \textbf{81.0} \\
\midrule
\rowcolor{rowhl} MiroThinker-1.7-mini & 75.4 & 78.4 & 76.9 \\
\rowcolor{rowhl} MiroThinker-1.7 & 76.5 & 78.5 & 77.5 \\
\rowcolor{rowhl} MiroThinker-H1 & \textbf{76.8} & 79.1 & 78.0 \\
\bottomrule
\end{tabular}
\end{table}

We next evaluate the ability of MiroThinker to generate high-quality long-form reports. 
Following the automated query generation framework of DeepResearchEval~\citep{wang2026deepresearcheval}, we construct a benchmark consisting of 50 deep research queries.  
We then compare MiroThinker with 10 representative deep research agents on these queries using the DeepResearchEval evaluation pipeline.
For each generated report, we evaluate two core dimensions: \textit{Report Quality} and \textit{Factuality}. 
Report Quality measures the overall quality of the generated report across multiple aspects, including coverage, insight, instruction-following, clarity, and task-specific evaluation dimensions. 
Factuality evaluates whether the statements in the report are accurate and grounded in evidence retrieved from the web.

Results are summarized in Table~\ref{tab:model-comparison-report}. 
Overall, the MiroThinker series demonstrates strong performance in long-form research report generation. 
We highlight two key findings. 
(a) \textit{State-of-the-art report quality.} MiroThinker-H1 achieves the highest report quality among the evaluated deep research agents, outperforming strong agents such as ChatGPT-5.4 Deep Research and Gemini-3.1-Pro Deep Research, indicating its strong capability in synthesizing complex information and producing high-quality long-form reports. 
(b) \textit{Strong factual grounding.} Our MiroThinker series surpasses most deep research agents and approaches the level of the strongest ChatGPT-5.4 deep research, demonstrating reliable factual grounding while generating comprehensive reports.

\subsection{Effective Interaction Scaling}

\begin{figure}[t]
    \centering
    \includegraphics[width=0.75\textwidth]{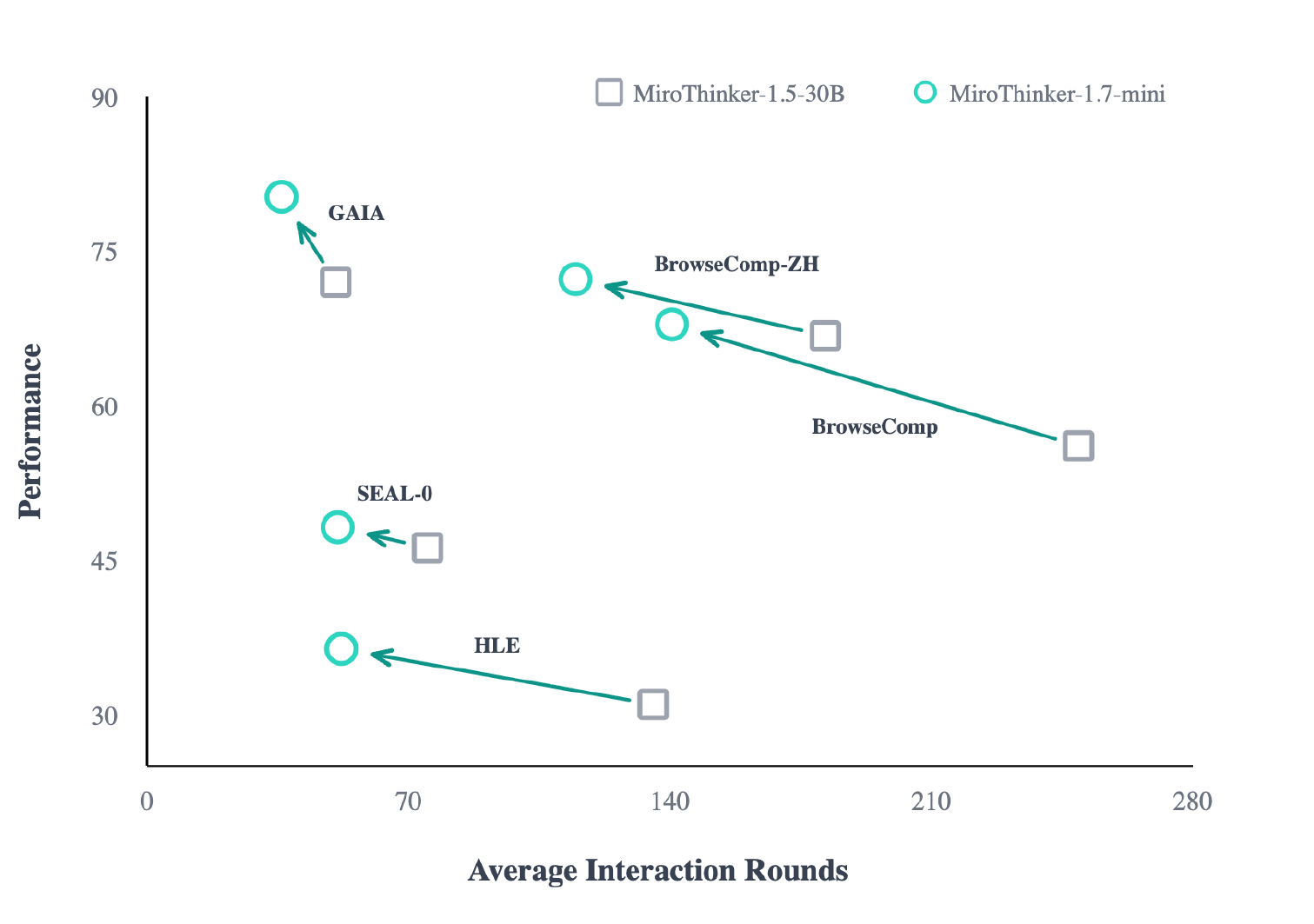}
    \caption{Performance vs. average interaction rounds. Arrows trace improvements from MiroThinker-1.5-30B to MiroThinker-1.7-mini (30B). All trajectories move upper-left, indicating higher performance with fewer turns.}
    \label{fig:performance_vs_turns}
\end{figure}

We argue that increasing the number of interaction turns does not necessarily translate into more effective interactions. When intermediate steps fail to produce meaningful progress toward solving the task, longer interaction trajectories may introduce redundant reasoning, propagate earlier mistakes, or increase exploration of unproductive paths. As a result, simply extending interaction length does not reliably improve task performance.

To examine this, we compare MiroThinker-1.5 and 1.7-mini under identical parameter budgets (30B) across five agentic benchmarks. As shown in Figure~\ref{fig:performance_vs_turns}, MiroThinker-1.7-mini consistently achieves higher performance with substantially fewer interaction rounds -- on average, 16.7\% better performance with about 43.0\% less rounds on the five benchmarks. The improvement is particularly pronounced on long-horizon tasks: HLE shows  17.4\% better performance with 61.6\% less rounds.
These results support our hypothesis that effective interaction scaling depends on improving the quality of each step rather than simply increasing trajectory length. The mid-training stage introduced in MiroThinker-1.7, which emphasizes planning, reasoning, and summarization, enables more reliable atomic actions, making each step more likely to advance the solution rather than accumulate noise.
 
\subsection{Verification-Centric Heavy-Duty Reasoning}
\label{sec:heavy_duty}

Here, we highlight the special contributions of the Local Verifier and the Global Verifier in MiroThinker-H1.

\begin{table}[htbp]
\centering
\caption{Local Verification only on BrowseComp hard subset (295 questions). Steps include all retries.}
\label{tab:local_verifier}
\begin{tabular}{lcccc}
\toprule
\textbf{Model} & \textbf{Pass@1} & \textbf{$\Delta$} & \textbf{Steps} & \textbf{$\Delta$} \\
\midrule
MiroThinker-1.7 & 32.1 & -- & 1185.2 & -- \\
MiroThinker-H1 w/ Local Verifier Only & 58.5 & \textcolor{teal}{+26.4} & 210.8 & \textcolor{teal}{-974.4} \\
\bottomrule
\end{tabular}
\end{table}

\textbf{Local Verifier Only.}
We select 295 questions from BrowseComp where MiroThinker-1.7 frequently fails as a hard subset. Results of MiroThinker-H1 w/ Local Verification only are reported in Table~\ref{tab:local_verifier}.

We observe two interesting findings: (a) \textbf{Steps.} MiroThinker-H1 reduces the number of interaction steps from 1185.2 to 210.8, roughly one-sixth of MiroThinker-1.7. This suggests that the Local Verifier improves the effectiveness of each interaction step, rather than relying on brute-force trial and error. Notably, this reduction is \emph{not} an explicit design objective, but a natural byproduct of local verification. (b) \textbf{Performance.} The improvement on this hard subset (+26.4) is more pronounced than on the full BrowseComp benchmark (+14.2), indicating that the Local Verifier is particularly effective at correcting erroneous reasoning paths in challenging scenarios.




\begin{figure}[t]
    \centering
    \includegraphics[width=0.75\textwidth]{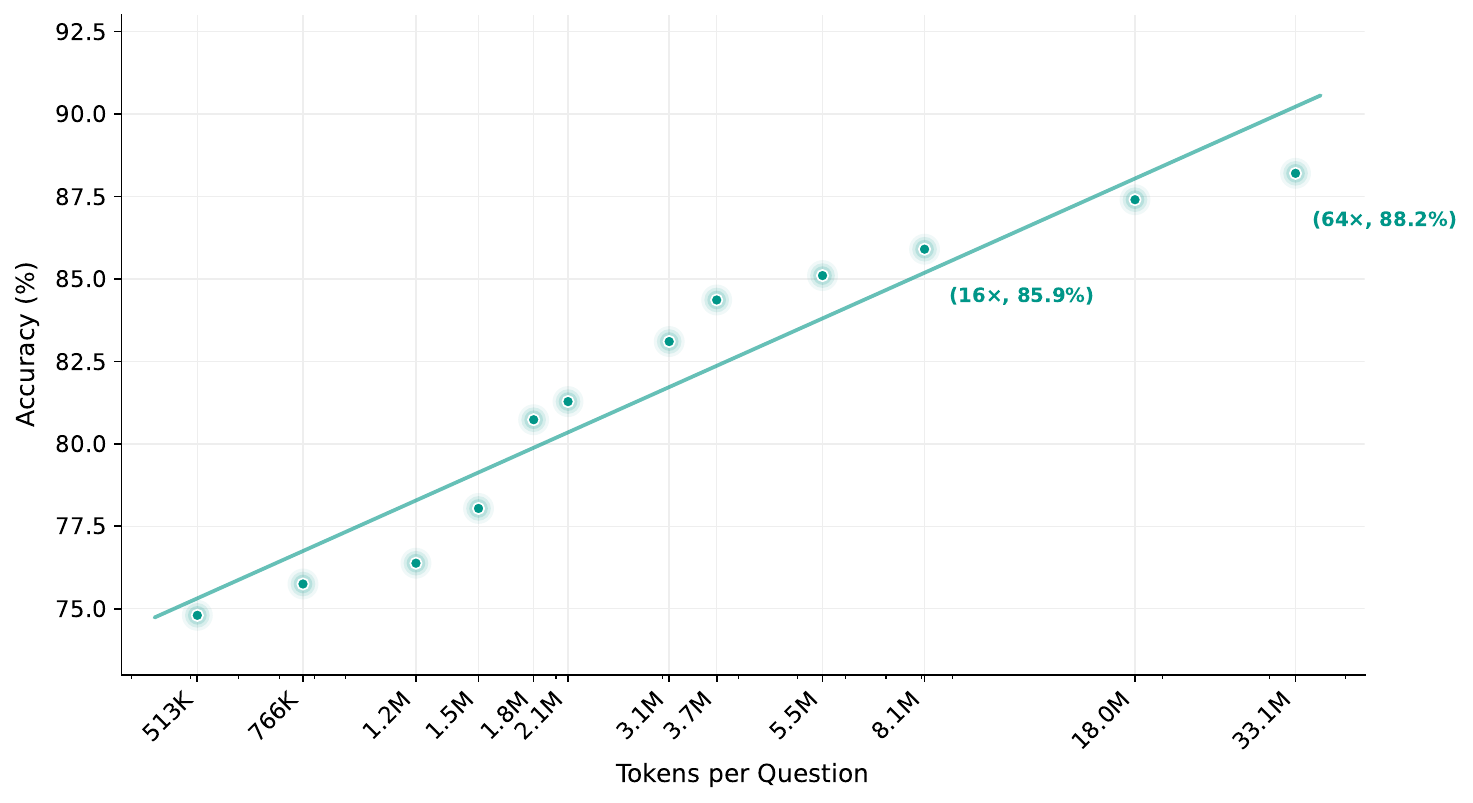}
    \caption{Token scaling curve of MiroThinker-H1 on BrowseComp. At 16× compute, the default budget for all benchmarks,       
  accuracy reaches 85.9. Scaling to 64× further improves accuracy to 88.2.}
    \label{fig:token_scaling_curve}
\end{figure}

\textbf{Global Verifier.} This module yields consistent improvements across all 
benchmarks, transforming MiroThinker-H1 into a heavy-duty system for search and reasoning tasks.

As shown in Table~\ref{tab:benchmarks-fast}, we highlight two noteworthy findings. (a) \textbf{Search-intensive tasks.} BrowseComp and Seal-0 achieve gains of +14.2 and +8.3 points respectively. These benchmarks require intensive web search or robust reasoning over noisy retrieval results, a setting where global verification provides the strongest advantage. As shown in Figure~\ref{fig:token_scaling_curve}, accuracy on BrowseComp scales log-linearly with 85.9, and scaling to 64$\times$ further improves accuracy to 88.2. (b) \textbf{Challenging reasoning tasks.} FrontierScience-Olympiad and HLE improve by 7.5 and 4.8 points respectively. These benchmarks demand complex reasoning coupled with accurate retrieval, indicating that global verification generalizes beyond search-intensive settings.

\section{Conclusions}


We introduce MiroThinker-1.7 and our flagship system, MiroThinker-H1, to address the inherent challenges of long-horizon reasoning in agentic AI. 
By emphasizing effective interaction scaling over mere trajectory lengthening, we developed an enhanced training pipeline that significantly improves planning, reasoning, and tool use capabilities. Furthermore, the integration of a heavy-duty, verification-centric reasoning mode at both local and global levels ensures that intermediate steps are continuously audited and refined before committing to a final solution. 
Extensive evaluations across diverse, complex benchmarks -- including BrowseComp, FrontierScience-Olympiad, and FinSearchComp -- demonstrate that MiroThinker-H1 establishes a new state-of-the-art, outperforming leading open-source and commercial research agents.
\bibliography{main}

\newpage

\section*{Contributions}

\subsection*{Core Contributors}
S. Bai, L. Bing, L. Lei, R. Li, X. Li, X. Lin, E. Min, L. Su, B. Wang, L. Wang, L. Wang, S. Wang, X. Wang, \\Y. Zhang, Z. Zhang

\subsection*{Contributors}
G. Chen, L. Chen, Z. Cheng, Y. Deng, Z. Huang, D. Ng, J. Ni, Q. Ren, X. Tang, B.L. Wang, H. Wang, N. Wang, C. Wei, Q. Wu, J. Xia, Y. Xiao, H. Xu, X. Xu, C. Xue, Z. Yang, Z. Yang, F. Ye, H. Ye, J. Yu, C. Zhang, \\W. Zhang, H. Zhao, P. Zhu  \\ \\

\section*{Acknowledgement}
We sincerely thank the following people who have left the team but previously made valuable contributions:

Carson Chen, Yuntao Chen, Zhe Chen, Jifeng Dai, Chenxia Han, Tammy Huang, Xiaoqi Jian, Shilei Jiang, Jerry Jiao, Ryan Luo, Ren Ma, Pax Sun, Hellen Wang, Weiyun Wang, Yan Xiao, Jinfan Xu, Enbo Zhao, Yanpeng Zhou, Xizhou Zhu




\end{document}